\documentclass[letterpaper, 10 pt, conference]{ieeeconf}  
\makeindex

\usepackage{graphicx}
\usepackage{multirow}
\usepackage{amsmath}
\usepackage[caption=false]{subfig}
\usepackage{cleveref}
\usepackage{amsmath}
\usepackage{mathtools}
\usepackage{afterpage}
\usepackage[table,xcdraw]{xcolor}
\usepackage{graphicx}

\usepackage{lipsum}
\usepackage{multicol}
\usepackage{graphicx}

\IEEEoverridecommandlockouts                              

\title{\LARGE \bf
DronePick: Object Picking and Delivery Teleoperation\\ with the Drone Controlled by a Wearable Tactile Display 
}

\author{Roman Ibrahimov$^{1}$, Evgeny Tsykunov$^{1}$, Vladimir Shirokun$^{1}$, Andrey Somov$^{2}$, and Dzmitry Tsetserukou$^{1}$
\thanks{$^{1}$Roman Ibrahimov, Evgeny Tsykunov, Vladimir Shirokun, and Dzmitry Tsetserukou are with the Intelligent Space Robotics Laboratory, Skolkovo Institute of Science and Technology, Moscow, Russian Federation.
        {\tt\small\{Roman.Ibrahimov, Evgeny.Tsykunov, Vladimir.Shirokun, Dzmitry. Tsetserukou\}@skoltech.ru}}%
\thanks{$^{2}$Andrey Somov is with Center for Computational and Data-Intensive Science and Engineering, Skolkovo Institute of Science and Technology, Moscow, Russian Federation.
        {\tt\small A.Somov@skoltech.ru}}%
}

\begin{document}

\maketitle
\thispagestyle{empty}
\pagestyle{empty}

\begin{abstract}
We report on the teleoperation system DronePick which provides remote object picking and delivery by a human-controlled quadcopter. The main novelty of the proposed system is that the human user continuously  gets the visual and haptic feedback  for accurate teleoperation. DronePick consists of a quadcopter  equipped with a magnetic grabber, a tactile glove with finger motion tracking sensor, hand tracking system, and the Virtual Reality (VR) application. The human operator teleoperates the quadcopter by changing the position of the hand. The proposed vibrotactile patterns representing the location of the remote object relative to the quadcopter are delivered to the glove. It helps the operator to determine when the quadcopter is right above the object. When the ``pick'' command is sent by clasping the hand in the glove, the quadcopter decreases its altitude and the magnetic grabber attaches the target object. The whole scenario is in parallel simulated in VR. The air flow from the quadcopter and the relative positions of VR objects help the operator to determine the exact position of the delivered object to be picked. The experiments showed that the vibrotactile patterns were recognized by the users at the high recognition rates: the  average 99\% recognition rate and the average 2.36s recognition time. The real-life implementation of DronePick featuring object picking and delivering to the human was developed and tested.
\vspace{10 mm}
\par
 
\end{abstract}

\section{Introduction}
One of the latest trends in Robotics is immersive teleoperation, when a human operator issues control commands to be remotely executed exactly as instructed. At the same time, operator receives rich feedback from the remote environment \cite{susumutachi2015}. The recent advancements in robot autonomy made it possible to eliminate the scenarios where human intervention is necessary -  the autonomous driving, for example. However, in a wide variety of domains, teleoperated human-controlled robots still play a crucial role. In robotic-assisted surgeries the high-definition view and advanced instruments assist the surgeon to diminish the size of  surgical wounds during the operation. Teleoperated robotics arms' ability to reach the tightest places that human hands cannot access makes medical surgeries much easier and less time consuming \cite{beasley2012medical, arbeille2003echographic}. Teleoperated robots are also widely used for safety, security, and rescue purposes \cite{birk2009networking,martins2009immersive}, such as deactivating bombs, monitoring undersea oil tubes, cleaning the radioactive and toxic waste. 
\par
 On the other hand, the main issue in robot teleoperation is the level of controllability of the system by the user which is dependent on the human-robot interaction. The real-life interaction between a human operator and a teleoperated robot poses a challenge for the human operator who is required to have agile operating and motor skills. Besides, some additional trainings are needed for the operator to be able to teleoperate a robot like a bomb disposal device. However, in such scenarios, the  weak interaction between the human and robot, as well as the lack of sufficient feedback from the remote side may lead to a failure of the mission. To solve such challenges, a number of approaches have been proposed to secure controll of the robots through joysticks, hand gestures, wearable devices, etc. All these methods have their own pros and cons.

\begin{figure}[t]
\centering
\includegraphics[width=\linewidth]{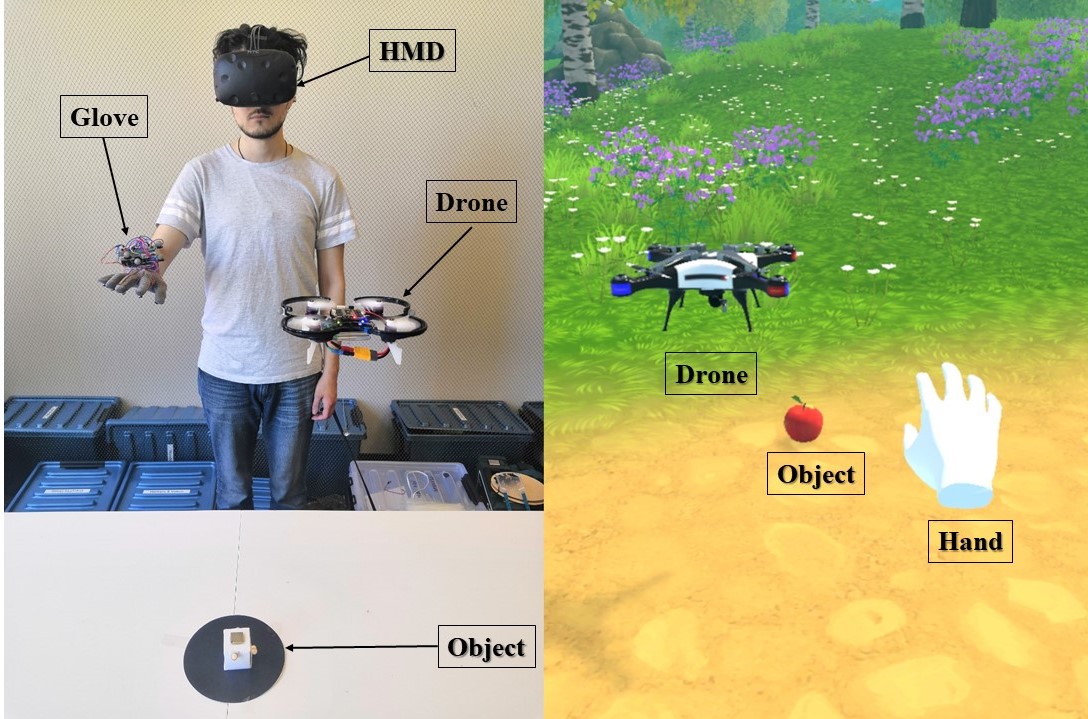}
\caption{A human operator manipulates the quadcopter to pick a remote object, (a) whereas he experiences it in VR (b).} 
\label{fig:EcUND} 
\end{figure}

\section{Related Work}
\subsection{Joystick-controlled Teleoperation}

The direct control of teleoperated devices and machines is ensured mainly by using joysticks, which help the users have straightforward guidance in the  remote hazardous areas. Even though recent research on the human-machine interface in teleoperation has resulted in new methodologies, such as EEG-based brain-computer interface \cite{christoforou2010android}, the majority of interfaces still rely on joystick control. When it comes to controlling the multi-robot system by a joystick, there is a bunch of well-known approaches. One of them is Virtual Structures where the entire formation of the swarm acts as a single entity and the geometric configuration during movement is kept stable \cite{lewis1997high, ren2004formation}. In \cite{zhou2018agile, zhou2015virtual}, a swarm of 3-5 quad-rotors was teleoperated by a single human operator using a standard gaming joystick, and each quad-rotor in the single-body swarm maintains a given state while avoiding collisions. Another teleoperated control system is proposed in \cite{cho2010teleoperation} where a mobile slave robot is controlled using a joystick from which the robot receives human operator's commands. The operator also receives the environmental information feedback through the force feedback. However, most of the control joysticks of teleoperated robots, such as quad-copters, are more complicated and require additional training for non-expert users. Within non-spacious environments in safety-critical scenarios, the experienced and trained personnel is needed for guiding quad-copters through the predetermined flight paths \cite{kosch2018dronectrl}.

\subsection{Hand Gesture-controlled Teleoperation}
In robotic teleoperations the human-machine interaction is usually provided via hand controllers, such as joysticks, sensor gloves, keyboards or mouse. Substituting these traditional interfaces with more natural ones has captured the attention of the researchers. The visual gesture recognition can be considered as one of the most natural interfaces as it is also used by human beings to interact with each other \cite{hu2003visual}. Hand gestures were widely adopted in the research of human-robot interaction, and there is a wide variety of approaches that have been proposed. In the gesture-based interface described by G. Podevijn et al. \cite{podevijn2013gesturing}, a swarm of ground robots gets commands from a human operator in the form of gestures that are captured by a Kinect system. In \cite{nagavalli2017multi}, human operators wear the  Myo armband which converts myoelectric muscle signals from hand gestures to control the input of the swarm of TurtleBots. In a similar scenario \cite{nagi2015wisdom}, a swarm of robots learns from the human hand gestures and, proceeding from the observations that each agent made, the swarm makes a unified decision. Although the above-mentioned works on human-robot interaction mentioned above have proven results, there are some limitations which should not be ignored. The gesture-based interaction requires the  complex infrastructure setup which narrows down the range of applications where it could be used. In those  scenarios where the system extracts the gestures from an image, the background of the user should be clear and straightforward so that the camera could easily identify gestures. The time spent on image acquisition and processing should be short enough to allow the system to function in real time \cite{chen2007human}.

\subsection{Wearable-controlled Teleoperation}

Haptics has also gained considerable attention in the research of human-robot interaction where various interaction methodologies have been proposed. Taking into account that the visual channel is overloaded in the process of teleoperation, the tactile interfaces deliver the feedback information about the swarm status directly to the skin. In \cite{tsykunov2018swarmtouch, labazanova2018swarmglove, tsykunov2019swarmtouch} the  authors propose the strategy where the formation of the swarm of nano-quadrotors with impedance control is guided by a human operator and the vibrotactile feedback maps the dynamic state of the formation through the tactile patterns at the fingertips. Likewise, in the robotic telepresence system \cite{tsetserukou2011belt} using the laser range finders (LIDARs) the mobile robot precisely recognizes the shape, boundaries, movement direction, speed, and distance to the obstacles. Then the tactile belt delivers the detected information to the user who regulates the movement direction and speed of the robot through the body stance (the operator's torso  works as a joystick). In another proposed system \cite{scheggi2014human}, the haptic bracelet which delivers the tactile feedback about the feasible guidance of a group of mobile robots in the motion constrained scenarios was developed. Although there might be some advantages of the haptic devices, such as getting a feedback from the places that are hard to obtain the visual feedback \cite{tsykunov2018swarmtouch}, these devices do not have a wide range of real-life applications due to a series of limitations. Haptics usually requires some  additional equipment to be set up, which in turn makes it difficult to be implemented in particular applications. In comparison with the  visual and audio feedback, the low bandwidth channel for the information transfer makes the tactile feedback less desirable. For example, the user might need to get the altitude status and obstacle warning simultaneously. In some highly demanding cases when a person cannot focus on his/her sensory input, the stimulus might not be felt \cite{LillyTechReport}.

\subsection{DronePick Approach}
The main novelty of the present paper is that we propose a system for the direct interactive teleoperation with a quadcopter by using VR and a tactile wearable to pick and deliver an intended object in remote environments. A new tactile glove  to be worn by the operator to control the quadcopter and to get information about the location of the remote object relative to the quadcopter is introduced. Four tactile patterns are also proposed for the glove. The teleoperated quadcopter, the remote environment and the remote object to be picked are simulated in the VR application. Unlike the solutions discussed earlier, in our approach, no additional advanced control equipment, special environment for the operation, piloting skills or high concentration are required for the human operators. Both visual feedback through the VR headset and vibro-tactile patterns through the wearable glove make it easy for the operator to precisely control the quadcopter, to pick a remote object, and to deliver it to the desired location. VR allows the operator to be located in a more pleasant environment. Additionally, VR helps concentrating on the most valuable details of the scene, i.e. the hand, the drone, and the object. This point secures the quality of teleoperation.   

\section{DronePick Technology}

\subsection{Drone with a Magnetic Grabber}

The flexibility and versatility of the Crazyflie 2.0 platform that help explore a wide range of research topics make it an ideal tool for the Robotics research. Although Crazyflie 2.0 can be programmed by a number of programming languages and additional small-sized sensory boards can be attached on it, its  small size of 92x92x29mm and the weight 27 grams limit the lifting of an additional weight. Therefore, in our approach, we extended the Crazyflie 2.0 platform to make a slightly bigger quadcopter with the  more thrust capability to lift the additional weight (see Fig.2). An external speed controller is attached to the Crazyflie 2.0 controller using BigQuad deck in order to control the more powerful brushless DC motors. The built quadcopter has the dimensions of 200x200x100 mm. 
\par
The control of the goal position of the quadcopter along the $X$ and $Y$ axes is as follows:
\begin{equation}   \label{eq:imp_z}
     \\ 
    x_{g}
    = 
    K\cdot\Delta x_{hum} 
    +
    x 
\end{equation}
\begin{equation}   \label{eq:imp_z}
     \\ 
    y_{g}
    = 
    K\cdot\Delta y_{hum} 
    +
    y
\end{equation}
where $x_{g}$ and $y_{g}$ are the goal positions, $K$ is the scaling coefficient which determines the effect of the human's hand movement, $\Delta x_{hum}$ and $\Delta y_{hum}$ determine the intensity of the operator's hand movement along each of the Cartesian axes with respect to the initial position, $x$ and $y$ are the previous coordinates of the quadcopter. 
\par
The position of the quadcopter along the $Z$ axis depends on the height of the glove and the ``clasp'' position of the controlling hand. When the operator lowers his/her hand below 1 m, the process stops and the quadcopter lands. If the operator clasps his/her hand, the flex sensor on the glove detects this gesture and forces the quadcopter to decrease its altitude to pick the object. The quadcopter returns back to the previous altitude when the operator opens his/her hand.

\subsection{Vibrotactile Wearable Glove}

\begin{figure}[t]
\centering
\includegraphics[width=0.49\textwidth]{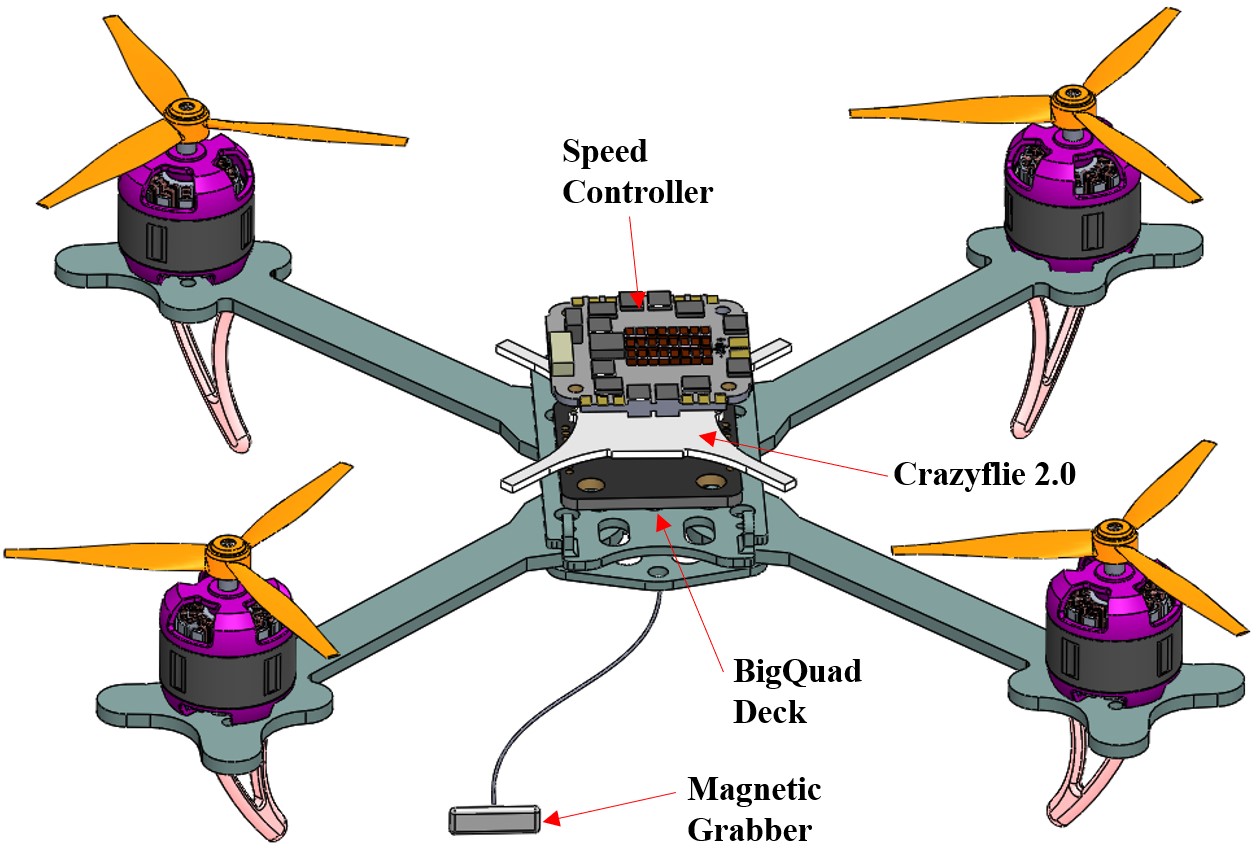}\label{true}
\caption{The Crazyflie 2.0 based quadcopter structure with a magnetic grabber.}
\label{swarmtouch}
\end{figure}

\begin{figure}[h!]
\centering
\subfloat[]{\includegraphics[width=1\linewidth]{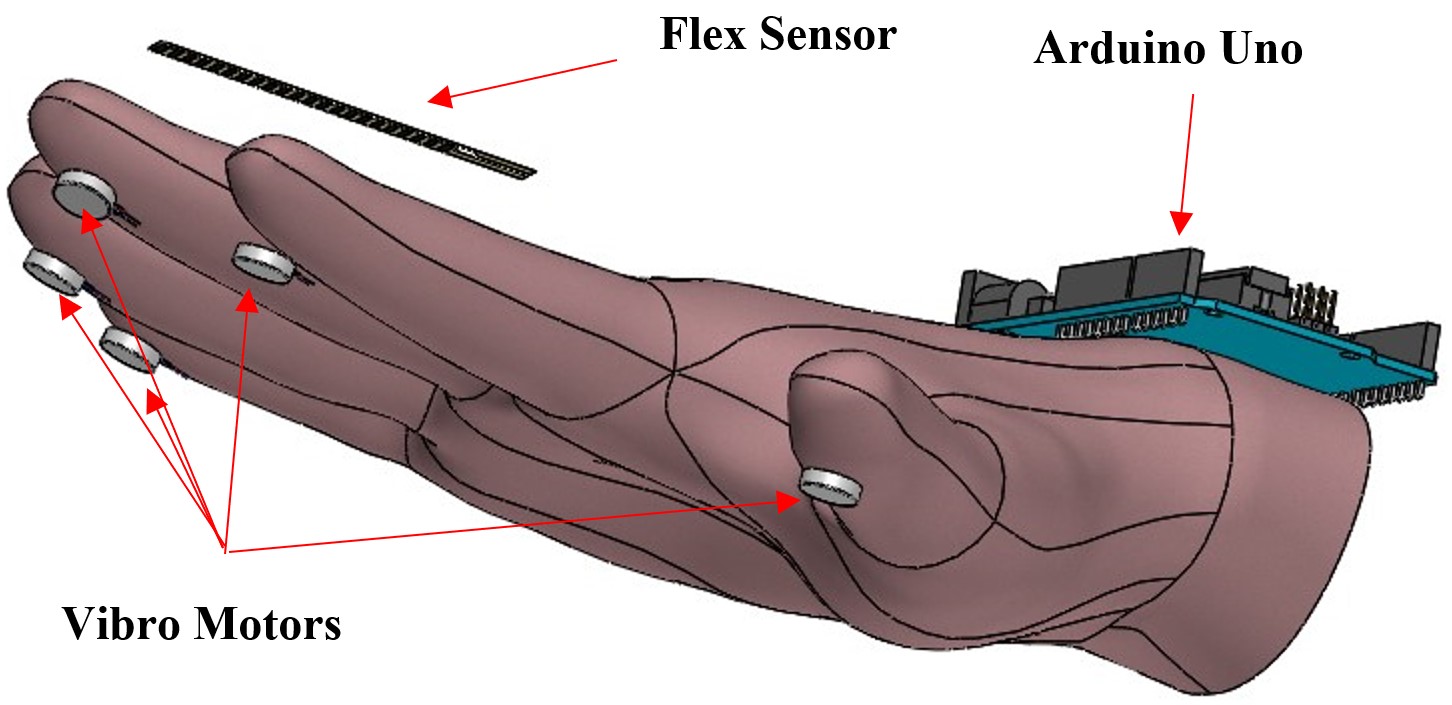}}

\label{fig:EcUND} 
\centering
\subfloat[]{\includegraphics[width=1\linewidth]{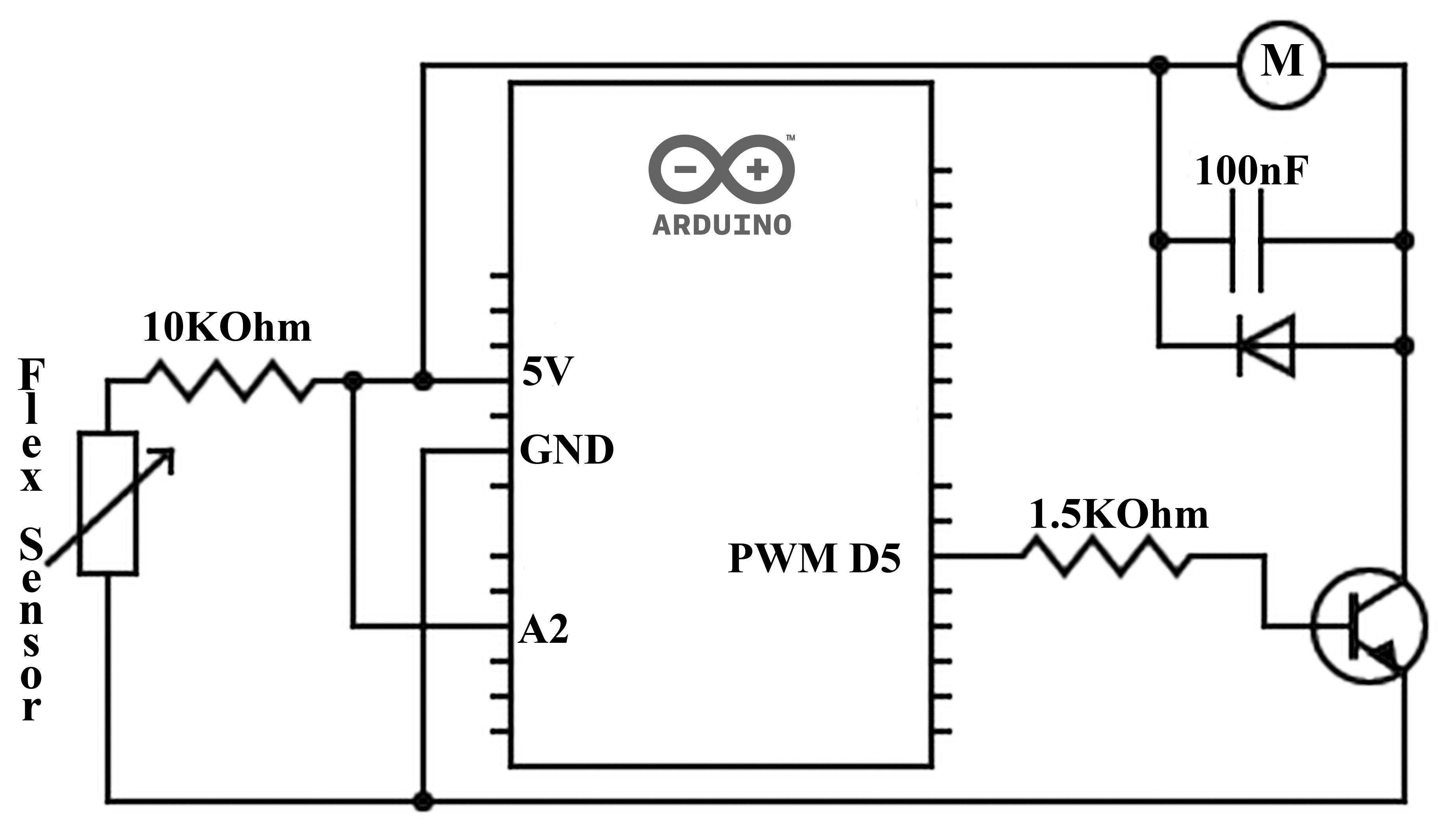}}

\caption{A tactile glove, (a) and its electrical circuit (b).}
\label{fig:EcUND} 
\end{figure}  

\begin{figure}[h!]
\centering
\includegraphics[width=0.49\textwidth]{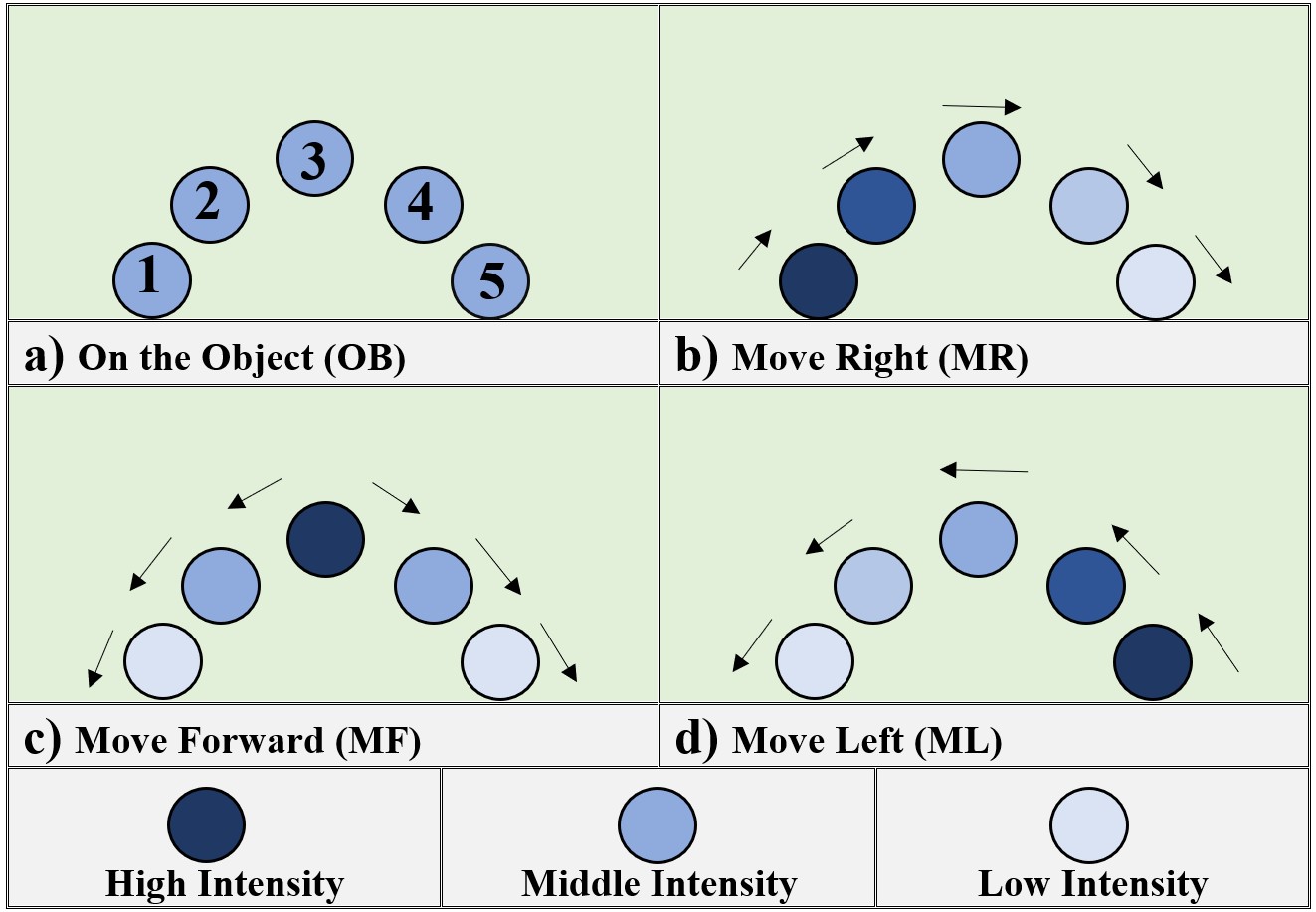}\label{true}
\caption{Tactile patterns for representing the remote object location. Each circle represents the right hand's finger (view from the dorsal side of the hand).}
\label{swarmtouch}
\end{figure}  
A human operator can receive essential information through the tactile wearables. The simple control and implementation of vibro-motors have made it possible to widely use them in tactile displays that reproduce the tactile sensation where the users can feel and touch virtual objects \cite{martinez2014identifying, maereg2017wearable}. We developed a unique wearable tactile glove with Eccentric Rotating Mass (ERM) vibro-motors at the fingertips which delivers  information about the position of the object relative to the quadcopter. The information is delivered to the operator in the form of tactile patterns. The motors receive control commands from the Arduino Uno controller which is connected to the computer with the  ROS master node running. The vibration motors are located at each fingertip to achieve the higher recognition rate of the patterns.

The vibration intensity of each of the motors individually changes based on the applied PWM signal to create different tactile patterns. One flex sensor is located along the middle finger. The clasping  position of the human hand is determined when the sensor is bent. It sends the ``pick'' command to the quadcopter whereas the quadcopter decreases its height to 15 cm above the ground until the hand is opened. Four Vicon Motion Capture markers are located on the upper side of the glove to determine the position of the glove in space with submillimeter accuracy. 
\par
The electrical circuit of the vibration motor control and the flex sensor reading are shown in Fig.3 (b). The diode and the capacitor are connected to the motor in parallel. The diode protects the circuit from the potential voltage spikes produced by the motor rotation whereas the capacitor absorbs the spikes. The transistor in the circuit functions as a switch which is activated by the PWM pulses from the digital port of Arduino Uno. The flex sensor can be considered as a variable resistor. Its resistance, which is read on the analog pin of the microcontroller, depends on the bending position of the middle finger. 
\par
The visual feedback is not always sufficient to precisely locate the quadcopter above the object to pick it. If, for instance, the object location has the same $X$ or $Y$ coordinate with respect to the  operator's location, it would be complicated for him/her to determine when the quadcopter is exactly  above the object. Therefore, four different tactile patterns were designed to deliver essential information about the object location to the operator (see Fig.4). ``On the Object (OB)'' tactile pattern with the same intensity level on each of the fingers  is played when the quadcopter is on the object. If the object is located on the right side of the drone, the  ``Move Right (MR)'' pattern with the highest intensity on the thumb and the lowest intensity on the little finger is played. Likewise, the operator receives the opposite pattern – ``Move Left (ML)'' when the object is located on the left side of the quadcopter. If the object is located in front of the drone, the ``Move Forward (MF)'' tactile pattern is played.

\subsection{Virtual Reality}
The VR application was created in Unity 3D engine to provide the user with the pleasant immersion while the  forest environment represents the remote location and target object to be picked is an apple. The quadcopter and the glove model were also simulated in VR using similar models. To optimize the CPU load and to decrease latency, the simplifying 3D model technology was used where the size of each virtual object changes as in real life depending on the distance between the operator and the objects. Vicon Motion Capture cameras were used to transfer the positioning and rotation tracking data of each object to the Unity 3D engine. The operator wears HTC Vive VR headset to experience the VR application. 

\begin{figure}[h!]
\centering
\includegraphics[width=0.49\textwidth]{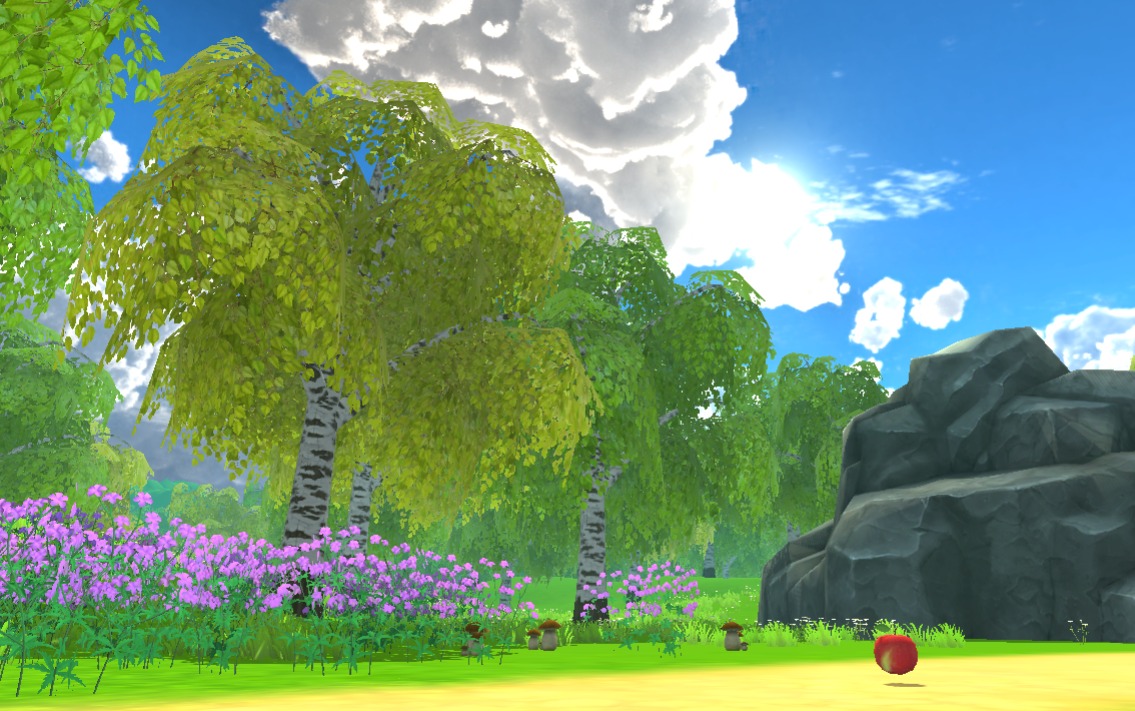}\label{true}
\caption{The intended object to be picked is shown as an apple in the VR application.}
\label{swarmtouch}
\end{figure}

\section{Experimental Results  and the Proof of Concept}
Ten right-handed volunteers, whose age range is between 22 and 33,  participated in the experiments. Five of them were male and the five were female. Before the start of experiment, each participant was given 5 minutes of training time to get used to and to familiarize with the tactile patterns. The volunteers positively responded to the wearable device convenience and pattern recognition level. \par 

The vibration intensity level of the tactile patterns varies from 100 Hz to 200 Hz. The highest intensity level is set 200 Hz, the middle intensity is to set 150 Hz, and the lowest intensity is set to 100 Hz (Fig.4).

\begin{table}[h!]
\centering
\caption{Confusion Matrix Showing the Recognition Rate}
\label{confusion_matrix}
\begin{tabular}{|p{1.2cm}|p{1.2cm}|p{1.2cm}|p{1.2cm}|p{1.2cm}|}
\hline
& \textbf{OB,} \% & \textbf{MR,} \%  & \textbf{MF,} \%  & \textbf{ML,} \%  \\ \hline
\textbf{OB,} \%  & \cellcolor[HTML]{76F7AE}\textbf{100.0 } & 0.0 & 0.0 & 0.0 \\ \hline
\textbf{MR,} \%  & 0.0 & \cellcolor[HTML]{76F7AE}\textbf{99.0} & 0.0 & \cellcolor[HTML]{FC839C}\textbf{1.0}  \\ \hline
\textbf{MF,} \%  & 0.0 & \cellcolor[HTML]{FC839C}\textbf{2.0} & \cellcolor[HTML]{76F7AE}\textbf{97.0} & \cellcolor[HTML]{FC839C}\textbf{1.0}  \\ \hline
\textbf{ML,} \%  & 0.0 &0.0 & 0.0& \cellcolor[HTML]{76F7AE}\textbf{100.0} \\ \hline

\end{tabular}
\end{table}
\par During the experiments, each pattern is played once before asking the volunteer to enter the pattern number. Overall, each pattern is played 10 times in random order. The results of the pattern recognition experiments are shown in Table 1. The green cells along the diagonal show the percentage of correct responses by the volunteers whereas the red cells indicate the percentage of incorrect responses that were mixed with other patterns. On average, the experiments showed that the recognition rate of the vibrotactile patterns is 99.0\%. Pattern ``On the Object (OB)'' and ``Move Left (ML)'' have full recognition rates whereas ``Move Right (MR)'' and ``Move Forward (ML)'' have 99.0\% and 97.0\% recognition rates correspondingly. 

\begin{table}[h!]
\centering
\caption{Average Time of Tactile Pattern Recognition}
\label{time_of_recognition}
\begin{tabular}{|p{1.2cm}|p{1.2cm}|p{1.2cm}|p{1.2cm}|p{1.2cm}|}
\hline
& \textbf{OB} & \textbf{MR} & \textbf{MF} & \textbf{ML}\\ \hline
\textbf{Time,} s & 1.86 & 2.36 & 2.83 & 2.39 \\ \hline
\end{tabular}
\end{table}

\begin{figure*}[h]
\includegraphics[width=\textwidth,height=12cm]{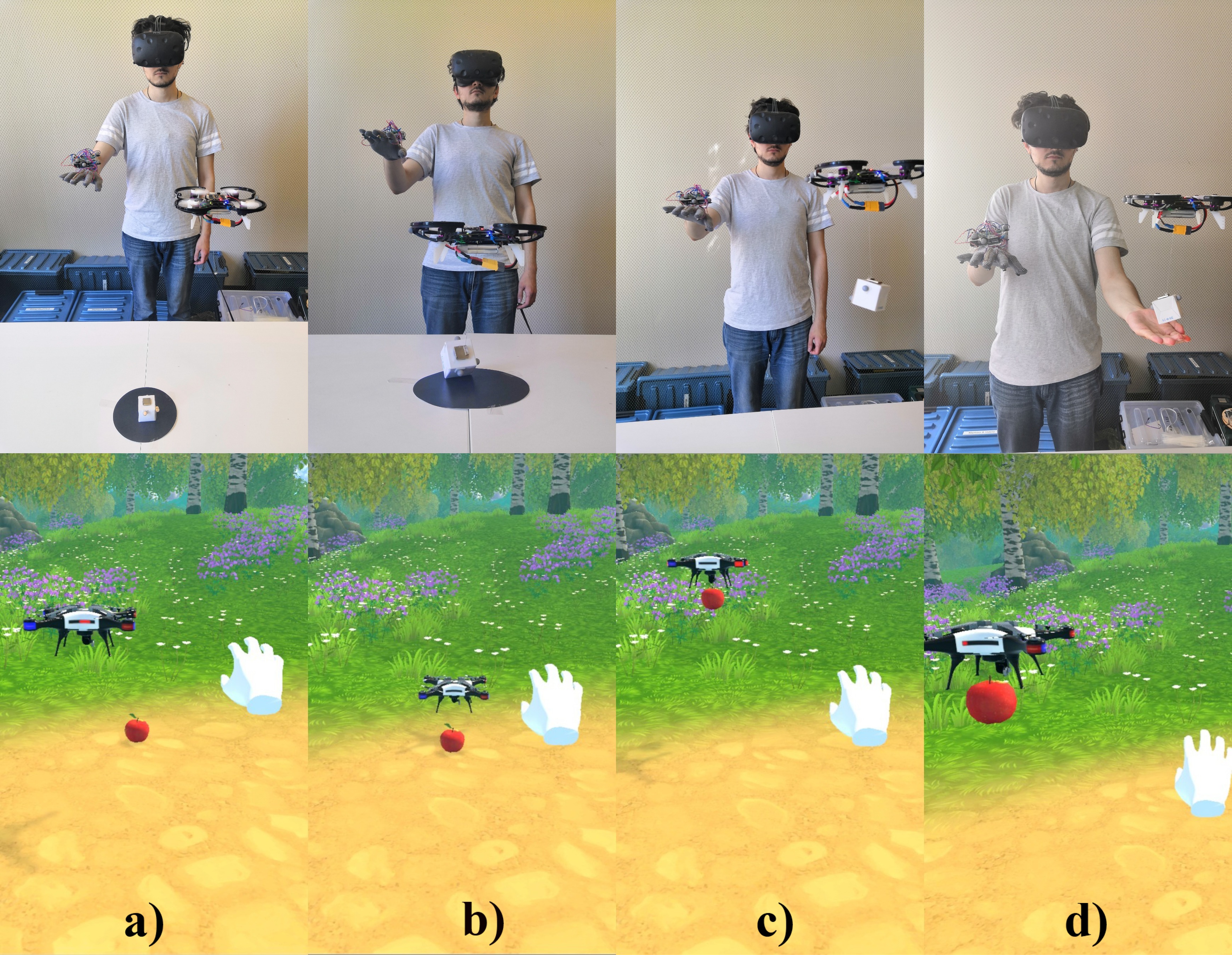}
\caption{A human operator: a) approaches the quadcopter to the object, b) picks the object by the quadcopter, c) delivers the object, d) picks the object from the quadcopter.}
\end{figure*}

During the experiments, the time spent on recognition patterns by the volunteers was also recorded. Recording started right after the pattern stopped playing and it finished when the user pressed the number button corresponding to the vibrotactile pattern. Average time spent on pattern recognition is separately displayed for each pattern in Table 2. The users overall spent the least time of 1.86s on ``On the Object (OB)'' pattern whereas 2.83s is the longest time interval that users spent on recognizing ``Move Forward (MF)'' pattern. The average recognition time for the patterns is 2.36 s. 

In addition to the tactile pattern recognition experiments, we implemented the real-life DronePick technology in the laboratory setting to prove the validity of the system. As it can be seen from  Fig.6, the human operator controls the quadcopter equipped with a magnetic grabber to pick and deliver the remote object. In parallel, he sees a simulated drone, a remote object as a red apple, and his hand with a glove in VR application. The picking and delivery process consists of four stages. Firstly, the operator approaches the quadcopter to the object (Fig.6 (a)). Tactile patterns helps him to locate it precisely right above the object. In Fig.6 (b), the object is being picked after ``pick" command is sent by clasping the hand in which a glove is worn. Fig.6 (c) shows how the quadcopter delivers the object toward the operator. Finally, in Fig.6 (d), the relative positions of the VR objects and the air flow from the quadcopter help the operator to determine the object and to pick it.

\section{Conclusion and Future Work}
In this work, we have proposed the novel system DronePick which combines the Virtual Reality (VR) and vibro-tactile wearable to provide the real-life  human-robot interaction, has been proposed. Both the  visual feedback in VR and the vibro-tactile feedback about the location of the object relative to the quadcopter bolster the human-robot interaction to  collaboratively fulfill object grabbing. Experiments on the pattern recognition showed the  average 99\% of recognition rate and the average 2.36 s recognition time. 
\par
The system might potentially have a substantial impact on the human-robot interaction in teleoperation. The DronePick technology can be applied to the  human-centric teleoperation scenarios where the  conventional ground robots are unable to perform. In the remote environments where the human presence is restricted, objects can be picked, removed or relocated by DronePick. 

\par
The future work provides for the complete user study  where the different user interfaces are tested to show the real impact of the new system onto the human-drone collaboration. The rate of reducing the cognitive and physical efforts for teleoperation of a drone with the aid of the DronePick technology is to be determined. Another experiment is to be the comparison of DronePick teleoperation with a) the glove and VR headset and with b) the glove only to demonstrate the  improved teleoperation by VR where the user is not in the field of operation. 

\par

To ensure easy grabbing of the object from the drone, the hand of the user can be visualized in a more precise way by using a tracking system, such as Leap Motion. The wireless and reliable communication between the glove and the ground station can be achieved by implementing the XBee radio or Wi-Fi modules. Furthermore, the system can be tested in the  Augmented Reality (AR) platform. This will allow the user to receive the more real information about the environment, and to reduce the risk of undesired contact with the drone's moving parts.

The scenario where human controls the swarm of drones to deliver heavy object to both hands can also be implemented. The proposed system merges the VR and real environment and makes the interaction tangible. Interestingly, the DronePick can find the application in the VR: with this device we do not need to render the haptic sensation of the object in the VR application since we can feel all physical properties (weight, texture, size, shape, temperature) from the real object delivered by the drone. These applications  of DronPick can considerably improve the realism of the VR experience.

\addtolength{\textheight}{-12cm}
\bibliographystyle{ieeetran}
\bibliography{bib}

\end{document}